\documentclass{article}

\usepackage[nonatbib, preprint]{neurips_2025}
\usepackage[numbers]{natbib}

\usepackage[utf8]{inputenc} 
\usepackage[T1]{fontenc}    
\usepackage{hyperref}       
\usepackage{url}            
\usepackage{booktabs}       
\usepackage{amsfonts}       
\usepackage{nicefrac}       
\usepackage{microtype}      
\usepackage{xcolor}         

\usepackage{booktabs}
\usepackage{colortbl}
\usepackage{multirow}
\usepackage{graphicx}
\usepackage[normalem]{ulem}
\useunder{\uline}{\ul}{}
\usepackage{subcaption}

\usepackage{amsmath, amssymb}
\usepackage[ruled,linesnumbered]{algorithm2e}

\usepackage{wrapfig}

\title{SAS-Bench: A Fine-Grained Benchmark for Evaluating Short Answer Scoring with Large Language Models}

\author{
  Peichao Lai$^1$ \And 
  Kexuan Zhang$^2$ \And 
  Yi Lin$^3$ \And 
  Linyihan Zhang$^2$ \And 
  Feiyang Ye$^2$ \And 
  Jinhao Yan$^2$ \And 
  Yanwei Xu$^{1}$ \And
  Conghui He$^{4}$ \And
  Yilei Wang$^2$ \And
  Wentao Zhang$^{1}$ \AND
  Bin Cui$^{1,*}$ \\
  $^1$Peking University \hspace{1em}
  $^2$Fuzhou University \\
  $^3$Hunan University \hspace{1em}
  $^4$Shanghai AI Laboratory \\
  $^*$\texttt{bin.cui@pku.edu.cn}
}

\begin{document}

\maketitle

\begin{abstract}
  Subjective Answer Grading (SAG) plays a crucial role in education, standardized testing, and automated assessment systems, particularly for evaluating short-form responses in Short Answer Scoring (SAS). However, existing approaches often produce coarse-grained scores and lack detailed reasoning. Although large language models (LLMs) have demonstrated potential as zero-shot evaluators, they remain susceptible to bias, inconsistencies with human judgment, and limited transparency in scoring decisions. To overcome these limitations, we introduce SAS-Bench, a benchmark specifically designed for LLM-based SAS tasks. SAS-Bench provides fine-grained, step-wise scoring, expert-annotated error categories, and a diverse range of question types derived from real-world subject-specific exams. This benchmark facilitates detailed evaluation of model reasoning processes and explainability. We also release an open-source dataset containing 1,030 questions and 4,109 student responses, each annotated by domain experts. Furthermore, we conduct comprehensive experiments with various LLMs, identifying major challenges in scoring science-related questions and highlighting the effectiveness of few-shot prompting in improving scoring accuracy. Our work offers valuable insights into the development of more robust, fair, and educationally meaningful LLM-based evaluation systems.
\end{abstract}

\section{Introduction}

The evaluation of Subjective Answer Grading (SAG) plays a vital role in education, standardized testing, and automated assessment systems. SAG tasks are generally categorized into two types: long-form, open-ended Automatic Essay Scoring (AES) \citep{DBLP:conf/ijcai/KhayiR24, DBLP:journals/corr/abs-2502-09497, DBLP:journals/air/RameshS22, ridley2020prompt, ridley2021automated} and short-form, close-ended Short Answer Scoring (SAS) \citep{wu2019short, wu2018short, menini2019automated}. The SAS task is typically formulated as a pointwise evaluation problem, where student responses are assessed against the reference answer, as is common in short-answer or proof-based questions. Traditional approaches to SAS have primarily focused on producing a single overall score for each response, often lacking fine-grained analysis. However, with recent advancements in large language models (LLMs) \citep{openai2023gpt4, deepseekai2024deepseekv3technicalreport}, novel paradigms have emerged in which LLMs serve as judges for zero-shot subjective scoring. These methods enable more comprehensive and nuanced assessments by incorporating multiple evaluation dimensions.

Despite the advantages of using LLMs as judges in zero-shot evaluation tasks, several challenges remain to be addressed. First, current LLMs exhibit notable biases in evaluation tasks, especially when compared to human assessors. These models are often influenced by factors such as the positioning of key phrases within the text and the overall length of the content \citep{DBLP:conf/emnlp/RainaLG24}, which can lead to biased scoring outcomes. Besides, LLMs tend to exhibit evaluation biases depending on the type of scoring metric used, whether it is a binary assessment (e.g., "yes" or "no") or a numerical scale (e.g., a 1–5 rating) \citep{DBLP:conf/naacl/ZhuangQHWYWB24}. Consequently, discrepancies between model-generated scores and human evaluations are significantly increased, leading to situations where the performance of LLMs in this task is even inferior to smaller language models (SLMs). Second, existing methods lack the capability to provide accurate and interpretable explanations for their evaluation results \citep{DBLP:journals/corr/abs-2412-14140}. For instance, these models fail to offer clear justifications for specific score assignments or to pinpoint the errors that contributed to inaccurate assessments, making manual verification more difficult. These limitations significantly hinder the practical deployment of LLM-based evaluation systems.

\begin{wrapfigure}
        {r}{0.33\textwidth} 
        \centering
        \includegraphics[width=0.33\textwidth]{
            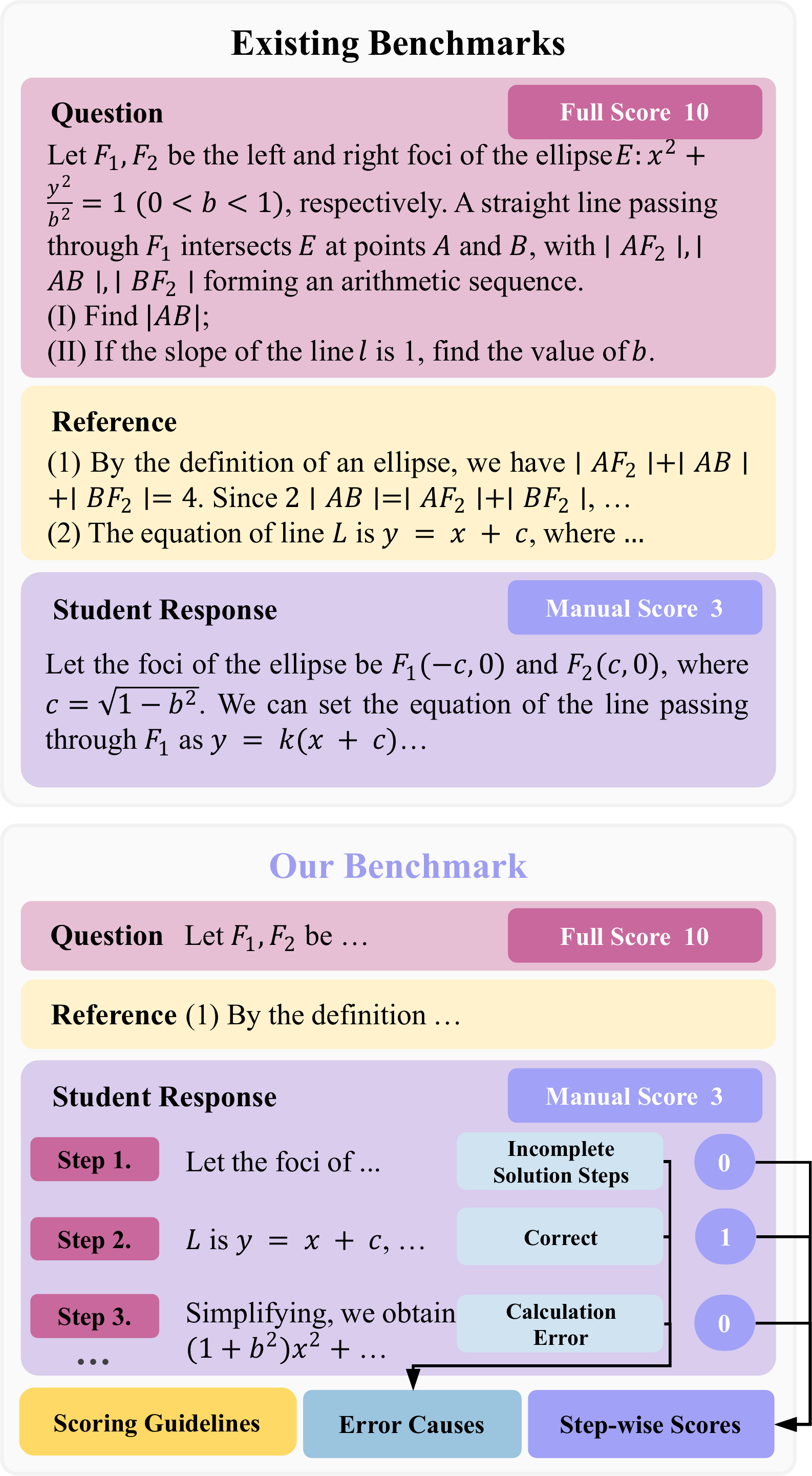
        } 
        \caption{Comparison of existing benchmarks and our benchmark.}
        \label{f1}
\end{wrapfigure}

To better understand how these challenges impact the SAS task, a benchmark specifically designed for generative language models is essential. As illustrated in Figure~\ref{f1}, most existing benchmarks rely on a simple input structure consisting of the question, reference answer, and student response, and evaluate model performance based on the alignment between predicted overall scores and human annotations. While this approach provides a general sense of scoring accuracy, it fails to capture the model's reasoning process and its ability to evaluate diverse forms of student responses. In particular, existing benchmarks still exhibit the following limitations: (1) Lack of Fine-Grained Evaluation. Most existing automatic scoring benchmarks focus on measuring the overall consistency between model-generated and human-assigned scores. However, they fail to capture how different response formats or key phrases within student answers may affect the model's scoring bias. Such fine-grained evaluation is particularly important for long-form responses, especially in science subjects, where accurate scoring often depends on specific reasoning steps and domain-specific terminology. (2) Limited Explainability Assessment. While automatic scoring models are bound to make occasional errors, their ability to interpret and justify their predictions is essential for enabling effective human oversight and review. Unfortunately, current benchmarks lack robust quantitative methods to assess the model's explainability. This shortcoming significantly hinders the adoption of these systems in high-stakes educational or assessment contexts.

To address these limitations, we present SAS-Bench, a new benchmark specifically designed for the SAS task. Built using authentic questions from China's National College Entrance Examination (Gaokao), the SAS-Bench dataset spans 9 academic subjects, comprising 1,030 questions and 4,109 student responses. All responses are manually annotated by subject-matter experts with step-wise scores and detailed error causes to ensure the accuracy and reliability of evaluation results. The benchmark also incorporates rigorous evaluation protocols to support consistent and fair comparison across models. \textbf{\textit{To support fine-grained assessment of model scoring performance}}, the dataset includes multiple template-free question types. Each student's response is segmented into multiple steps based on key phrases, allowing for detailed evaluation of how models handle different answer structures and reasoning processes. \textbf{\textit{To quantify model explainability}}, SAS-Bench incorporates a predefined set of error causes for each question type. Every step of each student's response is paired with expert-annotated error labels, enabling a systematic comparison between model-predicted and human-identified error causes. This facilitates a robust assessment of the model's ability to explain its scoring decisions. To evaluate the current capabilities of LLMs in SAS tasks, we conduct comprehensive experiments on sixteen widely used LLMs. Our results highlight that step-wise scoring and error causes inference remain particularly challenging in SAS tasks. Additionally, we find that providing few-shot demonstrations and scoring guidelines is positively correlated with improved model performance, offering valuable insights for enhancing LLM-as-a-Judger systems. In summary, our key contributions are as follows:

\begin{itemize}
\item \underline{\textit{New Benchmark.}} We present SAS-Bench, the first benchmark specifically tailored for SAS with LLMs. It features step-wise scoring and a predefined taxonomy of error causes, enabling more comprehensive evaluation of models' reasoning processes and the interpretability of their scoring decisions.

\item \underline{\textit{Accessible Dataset.}} We construct and publicly release an open-source dataset \footnote{https://github.com/PKU-DAIR/SAS-Bench} comprising 1,030 authentic questions with 4,109 student responses, each meticulously annotated by subject-matter experts with step-wise scores and detailed error labels, ensuring accurate and reliable evaluation.

\item \underline{\textit{Comprehensive Evaluation.}} We conduct extensive experiments on sixteen widely adopted LLMs, identifying major challenges in scoring short-answer responses and offering practical insights for enhancing LLM-as-a-Judger systems.
\end{itemize}

\section{Related Work}

\subsection{Automated Scoring Benchmarks}

Current automated scoring benchmarks can generally be categorized into two types: (1) Open-ended AES benchmarks without reference answers, such as the Kaggle ASAP-AES \citep{asap-aes}, ScAA \citep{DBLP:conf/icon-nlp/AgarwalGB20}, SciEntsBank \citep{DBLP:conf/semeval/DzikovskaNBLGBC13}, and the more recent RiceChem \citep{DBLP:conf/aied/SonkarNLKHB24}, which is tailored for evaluating LLMs. These datasets typically include question prompts, student essays, expert-assigned scores, and auxiliary metadata such as topic or user information. The recently introduced ENEM dataset \citep{silveira-etal-2024-new} further enriches evaluation by incorporating expert commentary on dimensions such as topic relevance and linguistic quality. (2) Close-ended SAS benchmarks with reference answers, which typically include a question prompt, a student response, and a corresponding reference answer. Representative datasets include the Kaggle ASAP-SAS \citep{asap-sas}, which spans multiple subjects including science, mathematics, and language arts; the SR dataset \citep{DBLP:conf/clic-it/MeniniTGV19}, which focuses on medical-domain questions; the ASAG dataset \citep{DBLP:conf/eacl/MohlerM09}, targeting short-answer questions in computer science; and the LE dataset \citep{DBLP:journals/csl/LaiYFCWW24}, which contains short-answer responses from the field of logistics management. The recently proposed L-Eval benchmark \citep{DBLP:conf/acl/AnG0ZLZKQ24} encompasses both open-ended and close-ended tasks, and introduces improved evaluation metrics for evaluating LLM grading performance on long-form responses. Despite their contributions, most existing datasets include only raw student responses and do not account for the influence of step-wise responses.


\subsection{LLM-as-a-Judge Systems}

With the rapid advancement of instruction-following capabilities in LLMs, the LLM-as-a-Judger paradigm has attracted growing interest. Numerous studies have explored prompting LLMs to directly assign assessment scores \citep{DBLP:journals/corr/abs-2303-04048, DBLP:conf/sigir/0001SC024, DBLP:conf/ictir/FaggioliDCDHHKK23, DBLP:conf/icmla/KaziK23}. However, as this line of research progresses, several limitations have been identified. Notably, LLMs exhibit high sensitivity to superficial variations in student responses, such as changes in the position or length of key phrases, which can significantly influence scoring behavior and make them vulnerable to adversarial attacks \citep{DBLP:conf/emnlp/RainaLG24}. In addition, LLMs often produce scoring outcomes that are inconsistent with human judgments, largely due to their susceptibility to variations in scoring rubrics or prompt wording \citep{DBLP:conf/naacl/ZhuangQHWYWB24}. To mitigate these challenges, recent work has proposed aligning LLM scoring with human evaluation standards by incorporating reference answers \citep{DBLP:conf/nips/ZhengC00WZL0LXZ23, DBLP:journals/corr/abs-2310-17631}, detailed scoring rubrics \citep{DBLP:conf/coling/LiuYHZHWDSZ24, DBLP:conf/bea/StahlBNW24, DBLP:conf/aied/LatifFMZ24}, and multi-perspective evaluation frameworks \citep{DBLP:conf/emnlp/LiusieRFG24}. These insights have informed the design of our benchmark, in which we provide explicit scoring guidelines and carefully curated reference answers to support a more rigorous, fair, and interpretable evaluation of LLM-based scoring systems.

\section{SAS-Bench}

\begin{figure}[h]
  \centering
  \includegraphics[width=\textwidth]{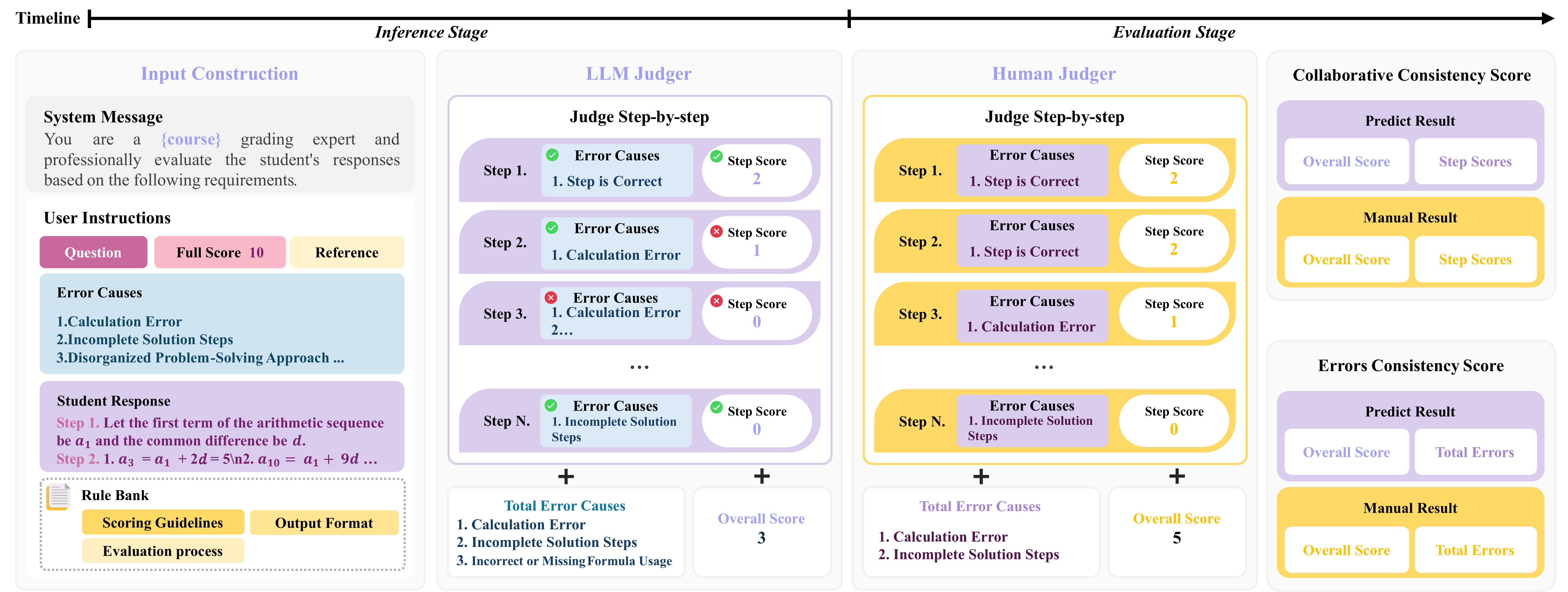}
  \caption{The workflow of our SAS-Bench. The results from the human judger are predefined during the dataset construction.}\label{f2}
\end{figure}

The overall workflow of SAS-Bench is illustrated in Figure~\ref{f2}. Section~\ref{s3.1} presents the underlying principles guiding our data design. Section~\ref{s3.2} details the dataset construction pipeline along with key statistics. Finally, Section~\ref{s3.3} introduces the evaluation methodology and defines the metrics used in our benchmark.

\subsection{Benchmark Design}\label{s3.1}

The motivation behind our benchmark design is driven by three key questions: 1) Can the LLM accurately assign overall scores across varying levels of student performance? 2) Can the LLM reliably evaluate step-wise scores for multi-step responses? 3) Can the LLM effectively identify and analyze the underlying causes of errors in student responses? To address these questions, we propose a LLM-as-a-Judger framework, in which the model is prompted via a system message to act as a subject-matter teacher. During the inference stage, each input instance includes the question, the full score, the corresponding reference answer, the associated set of error causes, the student's response, and a rule bank (containing scoring guidelines, output formats, etc.). Given this input, the model is tasked with generating the overall score, step-wise scores, and step-wise error causes for each response. In the evaluation stage, we assess the model's performance on the SAS task by comparing its judgments with human annotations across these three dimensions:

\textbf{Overall Score Consistency.} Overall score consistency is a core metric in the SAS task, as it reflects the model's ability to provide accurate holistic evaluations of student responses. By comparing model-predicted scores with human-assigned scores, we can assess how well the model aligns with expert judgment. The reliability of this metric directly impacts the credibility and utility of the automated scoring system.

\textbf{Step-wise Score Consistency.} While many existing SAS methods focus primarily on predicting overall scores, they often lack finer-grained evaluations. In practice, model-generated scores may not always be persuasive, particularly in complex multi-step problems. Evaluating the consistency between model-predicted and human-assigned step-wise scores offers a more detailed assessment of the model's scoring logic. Additionally, science subjects often involve multi-step reasoning, where the outcome of one step may influence subsequent steps. Step-wise score consistency thus serves as an indirect measure of the model's ability to capture inter-step dependencies and perform causal reasoning.

\textbf{Error Cause Consistency.} Explainability is a critical concern in SAS and related LLM-as-a-Judge tasks \citep{DBLP:journals/corr/abs-2412-14140, DBLP:conf/bea/StahlBNW24}. To address this, we propose explicitly predicting the underlying causes of errors in student responses. This not only enhances transparency in model decision-making but also allows us to assess the alignment between model interpretations and human judgments. Measuring the consistency of error cause predictions provides valuable insight into the model's interpretability and its potential to offer actionable feedback.

\subsection{Dataset}\label{s3.2}

Following the proposed design framework, the SAS-Bench dataset is developed through a structured pipeline comprising data synthesis, data cleaning, error cause set construction, and manual annotation. This process results in a high-quality dataset that incorporates multi-dimensional evaluation metrics.

\textbf{Data Synthesis.} To ensure that the question sources are authoritative and that the evaluation data covers a broad range of mainstream subjects, we adopt the GAOKAO-Benchmark \citep{DBLP:journals/corr/abs-2305-12474}, a publicly available dataset based on China's National College Entrance Examination (Gaokao), as our source of questions. To generate diverse answer data with varying score distributions and error types, we leverage LLMs to simulate student responses at different proficiency levels. Specifically, we first categorize the questions into three types: multiple-choice, gap filling, and short-answer questions. Although multiple-choice questions can be graded using rule-based methods, we include a small number in a template-free format to better assess LLMs' comprehension. We aim to synthesize positive and negative samples in a 1:1 ratio. Positive samples are created by prompting LLMs to either rewrite the reference answers in various styles or directly answer the questions. For negative samples, we design different construction strategies based on the question type: 1) For template-free multiple-choice questions, we first analyze the structure and number of correct options, randomly select distractors using a rule-based algorithm, and then prompt the LLM to express the selection in a free-form style. 2) For gap filling and short-answer questions, we randomly select fragments from the reference answers according to the target score distribution and instruct the LLM to introduce errors at specific points. 3) Additionally, for short-answer questions, we decompose the reference answers into multiple logical steps, randomly remove certain steps to create incomplete templates, and then employ a smaller-scale LLM to fill in the gaps. This approach generates more natural and less contrived incorrect responses.

\textbf{Data Cleaning.} For each question, we synthesize eight positive and eight negative samples. During the data cleaning phase, we focus on removing overly similar responses. Initially, we use GPT-4o to score the synthesized answers and group them based on the resulting score distribution. We then compute BLEU and Jaccard similarity scores between pairs of answers within each group, filtering out those with high redundancy. This process ensures that the final dataset preserves representative examples across different score levels while effectively eliminating synonymous paraphrases and heavily overlapping samples, thereby mitigating issues related to distribution imbalance.

\textbf{Error Causes Construction.} To evaluate the explainability of LLM judges, it is essential to assess their ability to identify and analyze the underlying causes of errors. However, quantitatively measuring this explainability is challenging, largely due to the diversity of unstructured responses and the difficulty of converting free-form explanations into a standardized numerical format. To address this challenge, we propose constructing a structured list of potential error causes for each subject and question type. By constraining LLMs to select error causes from this predefined list when explaining their scoring decisions, we enable a more systematic and quantifiable evaluation of model explainability. Specifically, for each subject and question type, we adopt a few-shot prompting strategy using representative questions, reference answers, and synthesized responses as input to GPT-4o, prompting it to generate 50 sets of error cause descriptions. Human annotators then consolidate and refine these descriptions, merging similar entries and simplifying the language to produce a comprehensive set of error causes that captures all major categories of mistakes.

\textbf{Human Annotation.} We recruited 18 subject-matter experts and divided them evenly into two groups: one for initial annotation and the other for verification. The first group annotated the data, while the second group reviewed and validated these annotations. In cases of significant discrepancies between the two groups' evaluations, the annotators engaged in discussions to reach a consensus, followed by re-annotation if necessary. During the annotation process, annotators were instructed to: (1) segment each response into distinct steps, (2) assign a score to each step, (3) label the error cause for each step, and (4) provide an overall score for the response. After completing the annotations for each subject and question type, annotators compiled a corresponding scoring guideline to support consistency in subsequent model-based evaluations.

\begin{figure}[t!]
  \centering
  \begin{minipage}{0.68\textwidth}
    \centering
    \captionof{table}{Statistics of our dataset. ``Avg. Steps'' and ``Max Steps'' denote the average and maximum number of steps per response, respectively. ``Avg. Length'' denotes the average length of per response.}
    \label{t1}
    \resizebox{\textwidth}{!}{%
        \begin{tabular}{ccccccccccc}
        \toprule
        \textbf{Question Type} & \textbf{Phy.} & \textbf{His.} & \textbf{Geo.} & \textbf{Bio.} & \textbf{Chi.} & \textbf{Math} & \textbf{Pol.} & \textbf{Eng.} & \textbf{Che.} & \textbf{Total} \\ \midrule
        Questions              & 47            & 128           & 28            & 116           & 181              & 412           & 60            & 49            & 9             & 1,030           \\
        Error Causes           & 10            & 5             & 5             & 7             & 9                & 9             & 4             & 9             & 7             & 65             \\
Avg. Steps             & 4.9          & 1.8          & 3.3          & 2.8          & 2.8             & 5.4          & 4.6          & 9.6          & 5.1          & 40.3          \\
Max Steps             & 16            & 8             & 10            & 8             & 12               & 25            & 12            & 16            & 8             & 115            \\
Responses              & 230           & 640           & 140           & 576           & 757              & 1,231          & 240           & 245           & 50            & 4,109          \\
Avg. Length             & 867.7            & 303.9             & 322.4            & 83.1             & 181.7               & 831.3            & 299.4            & 167.0            & 522.4             & 3578.9            \\ \bottomrule
        \end{tabular}%
        }
  \end{minipage}
  \hfill
  \begin{minipage}{0.3\textwidth}
    \centering
    \includegraphics[width=\linewidth]{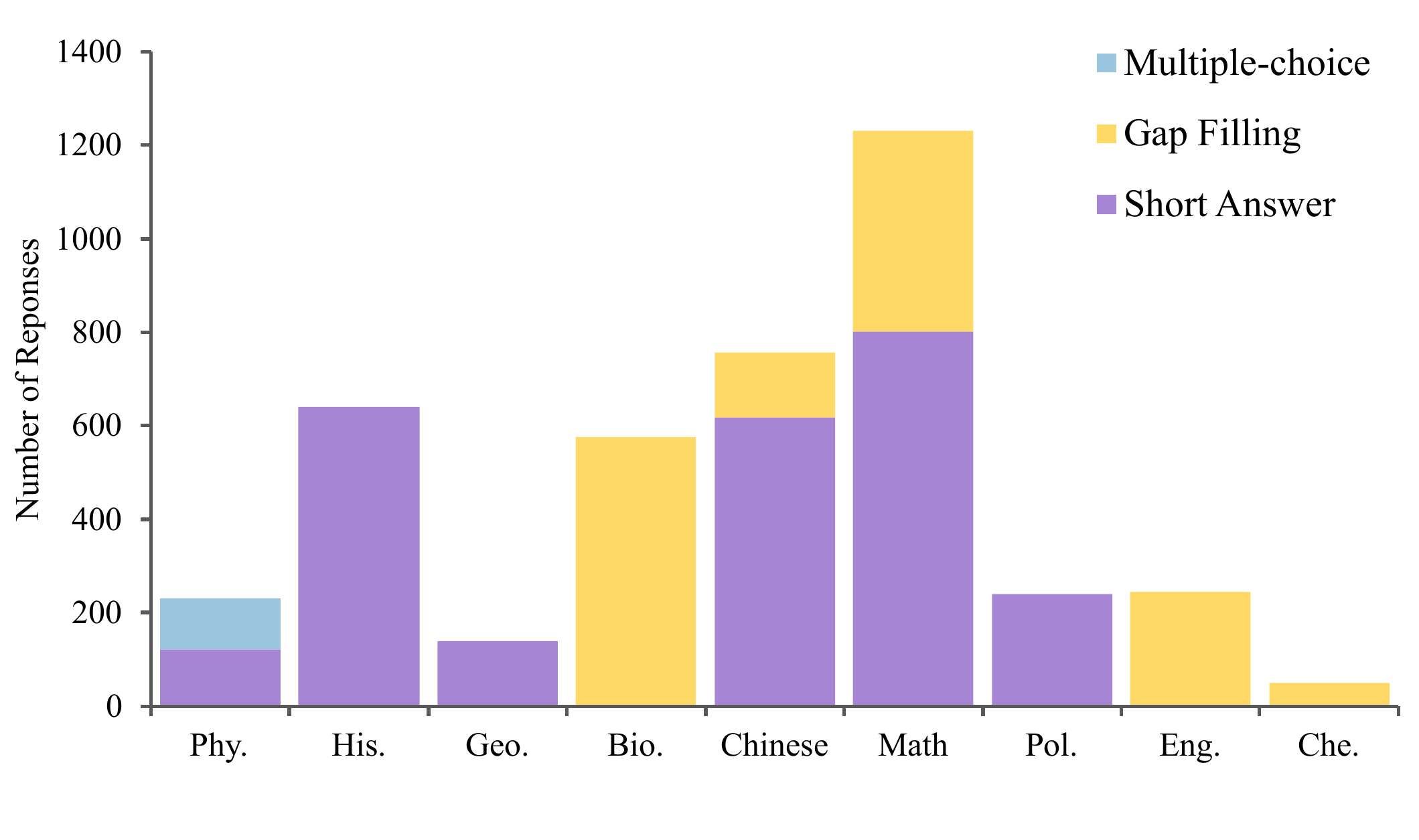}
    \caption{Distribution of question types across different subjects.}
    \label{f3}
  \end{minipage}
\end{figure}

\textbf{Data Statistics.} As shown in Table~\ref{t1}, the SAS-Bench dataset contains 4,109 student responses spanning nine subjects: Physics, History, Geography, Biology, Chinese, Mathematics, Politics, English, and Chemistry. The ratio of responses to questions is approximately 4:1. On average, each response includes six annotated error causes. Regarding the number of reasoning steps, science subjects tend to involve more detailed step-wise reasoning compared to humanities subjects. Notably, the Mathematics subject includes the response with the highest number of steps, reaching up to 25. Figure~\ref{f3} presents the distribution of question types. The dataset is primarily composed of short answer and gap filling questions. In addition to multiple-choice questions in Physics, the Chinese subject includes 440 responses with embedded multiple-choice items within short-answer questions.

\subsection{Evaluation Protocol}\label{s3.3}
Most existing benchmarks \citep{meyer2024asag2024, DBLP:conf/acl/MohlerBM11, DBLP:conf/semeval/DzikovskaNBLGBC13} provide only manually annotated overall scores, enabling model evaluation solely based on final scoring outcomes. However, they fall short in assessing step-wise scoring accuracy and the ability to identify specific error causes within responses. To more comprehensively evaluate LLM performance on this task, our benchmark simultaneously considers the alignment between model predictions and human annotations across three dimensions: overall scores, step-wise scores, and error cause interpretation.

While previous studies commonly adopt metrics such as Quadratic Weighted Kappa (QWK) \citep{DBLP:journals/corr/abs-2305-18638, DBLP:conf/ijcai/KhayiR24}, Pearson correlation \citep{DBLP:journals/corr/abs-2412-14140, DBLP:journals/eaai/BonthuSP23}, or Spearman correlation \citep{DBLP:conf/acl/LiuLWWWX0C023, DBLP:conf/naacl/ChuangDLZCS0YKG22} to assess scoring consistency, these metrics are primarily designed for evaluating overall scores and are not directly applicable to the multi-dimensional evaluation framework we propose. To overcome this limitation, we extend these traditional metrics and introduce two new evaluation measures tailored:

\textbf{Collaborative Consistency Score (CCS).} To jointly evaluate the consistency between model predictions and human annotations at both the overall and step-wise scoring levels, we propose the Collaborative Consistency Score (CCS). Formally, let \( r_i \) and \( r_j \) denote the holistic scores (e.g., overall scores) assigned by the model and human annotators, respectively, and let \( s_{i,k} \) and \( s_{j,k} \) denote their corresponding step-specific scores for the \( k \)-th step (\( k \in [1, m] \)). Then we define an adjusted weight matrix \( W \) that captures both holistic and step-wise discrepancies:
\begin{equation}
W_{i,j} = \alpha \cdot \frac{(r_i - r_j)^2}{(N_r - 1)^2} + \frac{1 - \alpha}{m} \sum_{k=1}^{m} \frac{(s_{i,k} - s_{j,k})^2}{(N_{s_k} - 1)^2},
\end{equation}
where \( \alpha \in [0, 1] \) controls the trade-off between holistic and step-wise differences (e.g., \( \alpha = 0.5 \)), \( N_r \) denotes the number of possible holistic score levels, \( N_{s_k} \) denotes the number of possible score levels for step \( k \), and \( m \) is the total number of steps. The CCS is then computed as:
\begin{equation}
\text{CCS} := 1 - \frac{\sum_{i,j} O_{i,j} \cdot W_{i,j}}{\sum_{i,j} E_{i,j} \cdot W_{i,j}},
\end{equation}
where \( O \) represents the observed matrix and \( E \) represents the expected matrix, both following the standard definitions used in the QWK metric.

\textbf{Errors Consistency Score (ECS).} To evaluate an LLM's ability to interpret error causes, we propose the Errors Consistency Score (ECS), which measures the consistency between model and human annotations in terms of error distributions. Although error causes are annotated at each step, we observe that both the number of steps and error types are relatively sparse, leading to instability in fine-grained step-wise consistency evaluation. Therefore, we aggregate error causes across all steps within a response, preserving both their types and frequencies. To further enhance stability, we partition the evaluation samples into score intervals based on their normalized overall scores and compute consistency scores within each interval. Specifically, let \( p \) and \( g \) denote the normalized overall scores predicted by the model and assigned by human annotators, respectively. Let \( \mathbf{E}^p, \mathbf{E}^g \in \mathbb{R}^{n \times l} \) represent the error frequency matrices for the model and human annotations, where \( n \) is the number of samples and \( l \) is the number of error types. Given \( m \) quantile thresholds \(\{\tau_q\}_{q=1}^{m-1}\) that divide the samples into \( m \) score intervals, we define the interval assignment function and the corresponding accumulated error frequencies as follows:
\begin{equation}
\phi(x) = \sum_{q=1}^{m-1} \mathbb{I}(x \geq \tau_q), \quad 
\mathbf{M}^p_k[j] = \sum_{i: \phi(p_i) = k} \mathbf{E}^p_{i,j}, \quad 
\mathbf{M}^g_k[j] = \sum_{i: \phi(g_i) = k} \mathbf{E}^g_{i,j},
\end{equation}
where \( \mathbb{I}(\cdot) \) is the indicator function, \( k \in \{0,1,\dots,m-1\} \) denotes the interval index, and \( j \in [1,l] \) indexes the error types. We then compute the Spearman rank correlation coefficient \( \rho_k \) between \(\mathbf{M}^p_k\) and \(\mathbf{M}^g_k\) within each interval, and the ECS is computed as the average correlation across all intervals:
\begin{equation}\label{eq_ecs}
\text{ECS} := \frac{1}{m} \sum_{k=0}^{m-1} \rho_k, \quad \rho_k = \text{SpearmanR}(\mathbf{M}^p_k, \mathbf{M}^g_k).
\end{equation}

\section{Experiments}\label{s4}

\subsection{Experimental Setup}\label{s4.1}

\begin{table}[b!]
    \caption{CCS score (\%) comparison across various LLMs, evaluated by subject area and question type. \textbf{S.} indicates Short Answer, \textbf{M.} denotes Multiple Choice, and \textbf{G.} refers to Gap Filling. The best and second-best scores in each category are highlighted in \textbf{bold} and \underline{underlined}, respectively.}
    \label{t2}
    \resizebox{\textwidth}{!}{%
    \begin{tabular}{lccccccccccccc}
    \toprule
    \textbf{Models}        & \textbf{\begin{tabular}[c]{@{}c@{}}Phy. (S.)\end{tabular}} & \textbf{\begin{tabular}[c]{@{}c@{}}Phy. (M.)\end{tabular}} & \textbf{\begin{tabular}[c]{@{}c@{}}His. (S.)\end{tabular}} & \textbf{\begin{tabular}[c]{@{}c@{}}Geo. (S.)\end{tabular}} & \textbf{\begin{tabular}[c]{@{}c@{}}Bio. (G.)\end{tabular}} & \textbf{\begin{tabular}[c]{@{}c@{}}Chi. (G.)\end{tabular}} & \textbf{\begin{tabular}[c]{@{}c@{}}Chi. (S.)\end{tabular}} & \textbf{\begin{tabular}[c]{@{}c@{}}Math (S.)\end{tabular}} & \textbf{\begin{tabular}[c]{@{}c@{}}Math (G.)\end{tabular}} & \textbf{\begin{tabular}[c]{@{}c@{}}Pol. (S.)\end{tabular}} & \textbf{\begin{tabular}[c]{@{}c@{}}Eng. (G.)\end{tabular}} & \textbf{\begin{tabular}[c]{@{}c@{}}Che. (G.)\end{tabular}} & \textbf{Avg.} \\ \midrule
    \multicolumn{14}{c}{Reasoning-based LLMs}                                                                                                                                                                                                                                                                                                                                                                                                                                                                                                                                                                                                                                                                                                                                                                                                                                                      \\ \midrule
Deepseek-R1            & 38.43                                                             & \textbf{95.01}                                                & \textbf{80.98}                                                    & 67.92                                                             & \textbf{79.12}                                                     & {\ul 95.09}                                                           & {\ul 69.07}                                                          & 57.85                                                             & \textbf{83.56}                                                     & 71.92                                                             & {\ul 73.19}                                                        & {\ul 72.92}                                                        & {\ul 73.76}    \\
QwQ-32B                & 48.53                                                             & {\ul 87.23}                                                   & 75.43                                                             & \textbf{77.06}                                                    & 72.52                                                              & \textbf{96.00}                                                        & 31.77                                                                & 48.66                                                             & 45.51                                                              & {\ul 74.48}                                                       & 54.79                                                              & 62.17                                                              & 64.51          \\
TinyR1-32B-Preview     & 38.17                                                             & 84.88                                                         & {\ul 75.83}                                                       & 71.52                                                             & 73.45                                                              & 92.57                                                                 & 52.61                                                                & 48.28                                                             & 74.77                                                              & 70.70                                                             & 57.92                                                              & 41.37                                                              & 65.17          \\
Qwen3-32B              & 47.29                                                             & 85.51                                                         & 64.96                                                             & 80.43                                                             & 63.15                                                              & 92.21                                                                 & 50.43                                                                & 51.26                                                             & {\ul 80.77}                                                        & 73.30                                                             & 59.33                                                              & 57.82                                                              & 67.20          \\
Qwen3-8B               & {\ul 54.33}                                                       & 76.17                                                         & 45.54                                                             & 68.89                                                             & 43.22                                                              & 86.01                                                                 & 42.02                                                                & 46.33                                                             & 73.33                                                              & 64.25                                                             & 50.55                                                              & 50.52                                                              & 58.43          \\
MiMo-7B-RL             & 52.77                                                             & 41.01                                                         & 61.33                                                             & 67.10                                                             & 35.93                                                              & 54.72                                                                 & 43.09                                                                & 38.09                                                             & 55.79                                                              & 36.78                                                             & 34.69                                                              & 31.05                                                              & 46.03          \\
Deepseek-Prover-V2-7B  & 22.59                                                             & 10.75                                                         & 2.92                                                              & 30.71                                                             & 50.63                                                              & 55.48                                                                 & 12.95                                                                & 0.87                                                              & 2.29                                                               & 10.44                                                             & 30.19                                                              & 28.76                                                              & 21.55          \\
DeepSeek-R1-Distill-7B & 33.71                                                             & 29.24                                                         & 50.92                                                             & 32.35                                                             & 52.18                                                              & 52.44                                                                 & 44.29                                                                & 29.52                                                             & 39.55                                                              & 53.77                                                             & 32.98                                                              & 34.27                                                              & 40.44          \\ \midrule
\multicolumn{14}{c}{RLHF-based LLMs}                                                                                                                                                                                                                                                                                                                                                                                                                                                                                                                                                                                                                                                                                                                                                                                                                                                           \\ \midrule
Deepseek-V3            & 53.89                                                             & 85.72                                                         & 69.85                                                             & 76.23                                                             & {\ul 76.51}                                                        & 93.42                                                                 & \textbf{69.49}                                                       & \textbf{58.81}                                                    & 80.18                                                              & \textbf{76.75}                                                    & \textbf{73.82}                                                     & \textbf{74.64}                                                     & \textbf{74.11} \\
GPT 4o-mini-20240718   & \textbf{58.90}                                                    & 81.19                                                         & 54.85                                                             & {\ul 76.59}                                                       & 65.39                                                              & 87.65                                                                 & 55.25                                                                & 43.56                                                             & 37.38                                                              & 63.44                                                             & 22.60                                                              & 55.98                                                              & 58.56          \\
Llama3.3-70B-Instruct  & 45.34                                                             & 70.03                                                         & 72.02                                                             & 72.51                                                             & 67.94                                                              & 85.30                                                                 & 35.83                                                                & {\ul 58.60}                                                       & 74.97                                                              & 63.68                                                             & 67.60                                                              & 38.94                                                              & 62.73          \\
Mixtral 8×7B-Instruct  & 30.78                                                             & 42.27                                                         & 33.43                                                             & 4.99                                                              & 44.45                                                              & 29.85                                                                 & 24.00                                                                & 26.73                                                             & 70.04                                                              & 43.92                                                             & 33.40                                                              & 42.05                                                              & 35.49          \\
Qwen2.5-32B-Instruct   & 40.53                                                             & 77.02                                                         & 62.34                                                             & 74.50                                                             & 72.07                                                              & 94.85                                                                 & 66.37                                                                & 50.08                                                             & 32.59                                                              & 64.09                                                             & 53.35                                                              & 62.87                                                              & 62.56          \\
Qwen2.5-14B-Instruct   & 53.76                                                             & 66.12                                                         & 60.96                                                             & 74.30                                                             & 67.50                                                              & 92.81                                                                 & 63.08                                                                & 43.28                                                             & 75.62                                                              & 62.03                                                             & 56.34                                                              & 57.53                                                              & 64.44          \\
GLM4-9B-Chat           & 45.62                                                             & 52.33                                                         & 36.81                                                             & 69.41                                                             & 39.19                                                              & 63.92                                                                 & 42.94                                                                & 35.50                                                             & 56.95                                                              & 54.83                                                             & 33.92                                                              & 30.79                                                              & 46.85          \\
Llama3-8B-Instruct     & 41.09                                                             & 35.10                                                         & 37.52                                                             & 31.29                                                             & 32.19                                                              & 38.13                                                                 & 32.89                                                                & 23.55                                                             & 62.43                                                              & 37.78                                                             & 31.68                                                              & 29.27                                                              & 36.08          \\ \bottomrule
\end{tabular}%
    }
\end{table}

In the experiments, we evaluate SAS-Bench using LLMs of varying scales and types. Our analysis centers on three core challenges in applying LLMs to SAS tasks: (1) the consistency of overall score predictions across student responses of different quality levels; (2) the alignment between overall scores and step-wise evaluations in multi-step responses; and (3) the reliability of LLMs in predicting the frequency of different error causes in student responses.

\textbf{Metrics.} We adopt the CCS and ECS metrics introduced in $\S$~\ref{s3.3} to evaluate, respectively, the consistency between overall and step-wise scores and the consistency of model predictions on the frequency of different error causes. Additionally, we use Quadratic Weighted Kappa (QWK) to assess overall scoring performance at a high level, and employ standard metrics such as the F1 score to support performance analysis in specific evaluation scenarios.

\textbf{Models.} We conduct experiments on 16 LLMs, which are broadly classified into two categories: RLHF-based models and reasoning-based models. The RLHF-based models include Deepseek-V3 \citep{deepseekai2024deepseekv3technicalreport}, GPT-4 \citep{openai2023gpt4}, the LLaMA family \citep{DBLP:journals/corr/abs-2407-21783}, Qwen-2.5 \citep{qwen2, qwen2.5}, GLM-4 \citep{glm2024chatglm}, Mixtral \citep{jiang2024mixtralexperts}, and Deepseek-Prover-V2 \citep{ren2025deepseekproverv2advancingformalmathematical}. The reasoning-based models comprise Deepseek-R1 \citep{deepseekai2025deepseekr1incentivizingreasoningcapability}, QwQ \citep{qwq32b}, Qwen-3 \citep{qwen3}, Tiny-R1 \citep{tinyr1proj}, and MiMo-RL \citep{xiaomi2025mimo}. Among these, LLaMA3-MetaMath \citep{yu2024metamath}, MiMo-RL, Deepseek-Prover-V2, and Tiny-R1 are proprietary models specifically optimized for mathematical problem solving.

\textbf{Settings.} In the data synthesis process described in $\S$~\ref{s3.2}, we use GPT-4o-mini and Deepseek-V3 as large-scale LLMs to generate synthetic data. For smaller-scale models, including GLM4-9B, Qwen2.5-14B, and LLaMA3-8B, we use them to complete predefined incomplete templates. To increase the diversity of the generated content, we introduce randomness by varying the temperature during the generation process. For evaluation, we evaluate all models on 4$\times$ NVIDIA A800 80G, and set the temperature of all reasoning-based models to 0.6 to optimize their reasoning capabilities. For RLHF-based models, we default to a temperature of 0.0. However, if a model produces outputs with invalid formatting, we re-run the generation at a temperature of 0.6 to obtain valid responses.

\subsection{Scoring Consistency Results}


We evaluate the consistency between overall and step-wise scores using the CCS metric, with results across different subjects and question types presented in Table~\ref{t2}. To highlight the distinction between CCS and traditional overall score consistency metrics, we additionally report each model's QWK performance in the radar charts of Figure~\ref{f4}. We also present the distribution of model-predicted overall scores in Appendix~\ref{a_pred}.

\begin{wrapfigure}
        {r}{0.36\textwidth} 
        \centering
        \includegraphics[width=0.3\textwidth]{
            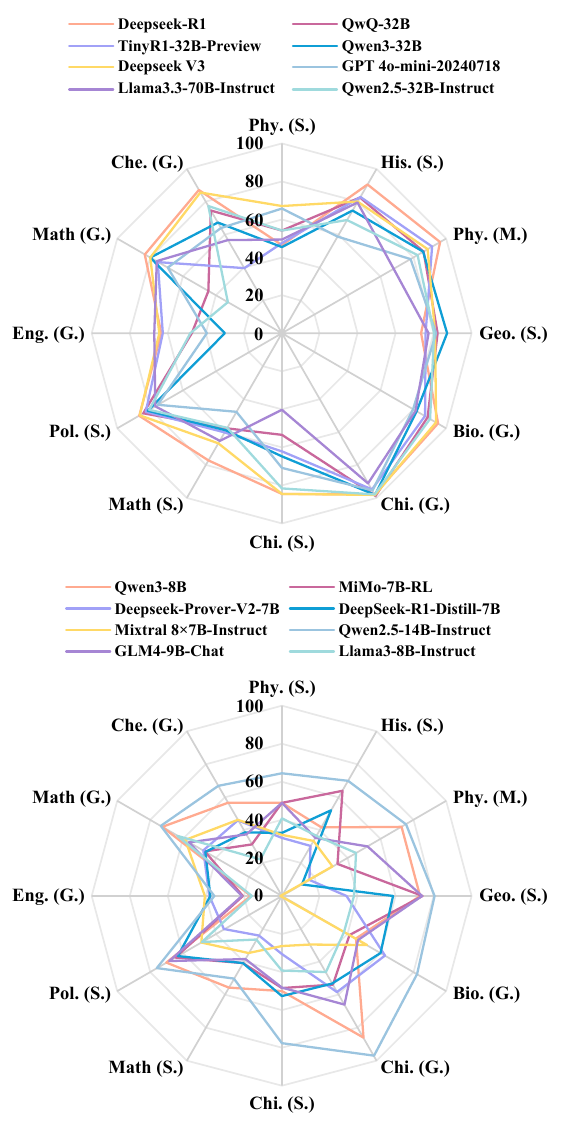
        } 
        \caption{Comparison of QWK scores across LLMs.}
        \label{f4}
\end{wrapfigure}

Overall, we observe a positive correlation between CCS and QWK scores across all models, as well as between model size and average performance. Notably, Deepseek-V3 and Deepseek-R1 achieve the best average results in CCS and QWK metrics, respectively. A comparison of Table~\ref{t2} and Figure~\ref{f4} shows that CCS scores are generally lower than QWK scores, indicating that incorporating step-wise consistency introduces additional challenges for scoring models. Models tend to perform better on humanities-related questions in both metrics, while science-related questions remain more difficult. For larger models with over 32B parameters, such as LLaMA3.3-70B-Instruct and Qwen3-32B, the CCS scores on short-answer questions like \textbf{His. (S.)} and \textbf{Chi. (S.)} are notably lower than their corresponding QWK scores. This discrepancy is more pronounced in reasoning-based models than in RLHF-based models. For instance, TinyR1-32B exhibits an 11.89\% drop in CCS compared to QWK on \textbf{Math (S.)}, suggesting that many LLMs still struggle with step-wise scoring consistency. Conversely, there are also counterexamples. MiMo-7B-RL, for example, achieves CCS scores on \textbf{Phy. (M.)}, \textbf{Phy. (S.)}, and \textbf{Eng. (G.)} that are, on average, 11.47\% higher than its QWK scores. This demonstrates that CCS provides a valuable and complementary perspective for evaluating model performance in SAS tasks. Finally, smaller-scale models (with fewer than 32B parameters) exhibit noticeable weaknesses in gap-filling tasks (\textbf{Eng. (G.)}) and template-free multiple-choice questions (\textbf{Phy. (M.)}), suggesting limited ability to handle sparse semantic content.

Building on the above experimental findings, the CCS metric offers more fine-grained insights into model performance on SAS tasks. By introducing step-wise consistency evaluation, CCS can guide future work toward enhancing step-by-step reasoning, thereby improving the practical utility of models in real-world scenarios.

\subsection{Errors Consistency Results}
\label{s4.3}

\begin{table}[h!]
    \caption{ECS score (\%) comparison across various LLMs, evaluated by subject area and question type. \textbf{S.} indicates Short Answer, \textbf{M.} denotes Multiple Choice, and \textbf{G.} refers to Gap Filling. The best and second-best scores in each category are highlighted in \textbf{bold} and \underline{underlined}, respectively.}
    \label{t3}
    \resizebox{\textwidth}{!}{%
    \begin{tabular}{lccccccccccccc}
    \toprule
    \textbf{Models}        & \textbf{\begin{tabular}[c]{@{}c@{}}Phy. (S.)\end{tabular}} & \textbf{\begin{tabular}[c]{@{}c@{}}Phy. (M.)\end{tabular}} & \textbf{\begin{tabular}[c]{@{}c@{}}His. (S.)\end{tabular}} & \textbf{\begin{tabular}[c]{@{}c@{}}Geo. (S.)\end{tabular}} & \textbf{\begin{tabular}[c]{@{}c@{}}Bio. (G.)\end{tabular}} & \textbf{\begin{tabular}[c]{@{}c@{}}Chi. (G.)\end{tabular}} & \textbf{\begin{tabular}[c]{@{}c@{}}Chi. (S.)\end{tabular}} & \textbf{\begin{tabular}[c]{@{}c@{}}Math (S.)\end{tabular}} & \textbf{\begin{tabular}[c]{@{}c@{}}Math (G.)\end{tabular}} & \textbf{\begin{tabular}[c]{@{}c@{}}Pol. (S.)\end{tabular}} & \textbf{\begin{tabular}[c]{@{}c@{}}Eng. (G.)\end{tabular}} & \textbf{\begin{tabular}[c]{@{}c@{}}Che. (G.)\end{tabular}} & \textbf{Avg.} \\ \midrule
    \multicolumn{14}{c}{Reasoning-based LLMs}                                                                                                                                                                                                                                                                                                                                                                                                                                                                                                                                                                                                                                                                                                                                                                                                                                                      \\ \midrule
Deepseek-R1            & 23.25                                                             & 30.59                                                         & 57.53                                                             & 56.08                                                             & 69.20                                                              & 86.04                                                                 & 72.68                                                                & \textbf{94.29}                                                    & 15.20                                                              & 65.56                                                             & {\ul 18.65}                                                        & {\ul 81.76}                                                        & \textbf{55.90} \\
QwQ-32B                & 4.74                                                              & \textbf{63.92}                                                & 67.06                                                             & {\ul 70.04}                                                       & 53.68                                                              & 51.08                                                                 & 69.20                                                                & 79.05                                                             & 16.82                                                              & 48.81                                                             & -22.53                                                             & 48.94                                                              & 45.90          \\
TinyR1-32B-Preview     & 3.10                                                              & \textbf{63.92}                                                & 65.71                                                             & \textbf{77.02}                                                    & 56.61                                                              & 64.42                                                                 & 74.83                                                                & 82.86                                                             & 23.33                                                              & 40.17                                                             & -31.52                                                             & 17.35                                                              & 44.82          \\
Qwen3-32B              & -4.17                                                             & 24.18                                                         & {\ul 69.52}                                                       & 54.29                                                             & 53.67                                                              & 52.70                                                                 & 47.31                                                                & 82.21                                                             & 18.33                                                              & 62.14                                                             & -26.99                                                             & 36.27                                                              & 39.12          \\
Qwen3-8B               & 23.39                                                             & \textbf{63.92}                                                & 14.29                                                             & -4.96                                                             & 52.21                                                              & 47.75                                                                 & 34.01                                                                & 39.20                                                             & -8.14                                                              & 57.19                                                             & -27.13                                                             & 59.28                                                              & 29.25          \\
MiMo-7B-RL             & \textbf{51.05}                                                    & 24.18                                                         & 14.29                                                             & 38.85                                                             & 58.35                                                              & {\ul 92.17}                                                           & 63.07                                                                & 13.39                                                             & 35.12                                                              & -27.10                                                            & -4.41                                                              & 1.04                                                               & 30.00          \\
Deepseek-Prover-V2-7B  & -24.10                                                            & -5.20                                                         & 42.86                                                             & -6.23                                                             & 29.54                                                              & -80.81                                                                & 23.25                                                                & 46.67                                                             & -1.51                                                              & -58.64                                                            & -45.23                                                             & -21.91                                                             & -8.44          \\
DeepSeek-R1-Distill-7B & -45.19                                                            & 24.18                                                         & 0.95                                                              & -38.66                                                            & 23.55                                                              & -20.36                                                                & 3.87                                                                 & -23.81                                                            & -13.57                                                             & -18.81                                                            & -19.59                                                             & -44.58                                                             & -14.34         \\ \midrule
\multicolumn{14}{c}{RLHF-based LLMs}                                                                                                                                                                                                                                                                                                                                                                                                                                                                                                                                                                                                                                                                                                                                                                                                                                                           \\ \midrule
Deepseek-V3            & 7.79                                                              & 46.58                                                         & 58.10                                                             & 32.62                                                             & {\ul 72.38}                                                        & \textbf{96.58}                                                        & 57.43                                                                & {\ul 92.38}                                                       & {\ul 33.33}                                                        & 40.26                                                             & \textbf{24.77}                                                     & \textbf{85.83}                                                     & {\ul 54.00}    \\
GPT 4o-mini-20240718   & 17.91                                                             & 24.18                                                         & 62.14                                                             & 36.68                                                             & 55.20                                                              & 79.01                                                                 & \textbf{78.00}                                                       & 67.62                                                             & \textbf{46.90}                                                     & \textbf{92.31}                                                    & 10.04                                                              & 36.39                                                              & 50.53          \\
Llama3.3-70B-Instruct  & 22.56                                                             & {\ul 57.35}                                                   & 54.29                                                             & 42.11                                                             & 45.09                                                              & 52.70                                                                 & 46.25                                                                & 54.29                                                             & 30.00                                                              & 58.81                                                             & -12.53                                                             & -15.83                                                             & 36.26          \\
Mixtral 8×7B-Instruct  & 11.99                                                             & 17.34                                                         & \textbf{80.38}                                                    & 35.84                                                             & 32.74                                                              & 42.77                                                                 & 75.82                                                                & 56.19                                                             & 30.00                                                              & 6.84                                                              & -31.16                                                             & -7.18                                                              & 29.30          \\
Qwen2.5-32B-Instruct   & 11.95                                                             & 17.41                                                         & 53.33                                                             & 59.34                                                             & 62.96                                                              & 46.90                                                                 & 75.08                                                                & 62.86                                                             & 30.00                                                              & 46.67                                                             & -4.50                                                              & 27.08                                                              & 40.76          \\
Qwen2.5-14B-Instruct   & 21.50                                                             & 24.18                                                         & 47.92                                                             & 37.43                                                             & \textbf{73.36}                                                     & 64.97                                                                 & 74.32                                                                & 64.94                                                             & 18.21                                                              & 61.97                                                             & -20.00                                                             & 47.39                                                              & 43.02          \\
GLM4-9B-Chat           & 35.00                                                             & 24.18                                                         & 32.49                                                             & 34.73                                                             & 62.12                                                              & 20.36                                                                 & {\ul 77.34}                                                          & 63.81                                                             & \textbf{46.90}                                                     & {\ul 82.40}                                                       & -25.35                                                             & 7.18                                                               & 38.43          \\
Llama3-8B-Instruct     & {\ul 48.25}                                                       & 27.46                                                         & 17.23                                                             & 31.58                                                             & 61.37                                                              & -14.05                                                                & 41.23                                                                & 57.77                                                             & 21.55                                                              & -69.07                                                            & -26.50                                                             & -27.19                                                             & 14.14          \\ \bottomrule
\end{tabular}%
    }
\end{table}

We employ the ECS metric to assess the consistency between model-predicted error cause frequencies and human annotations across different score levels. Specifically, we set $m$ to 3 in Equation~\ref{eq_ecs} to represent low, medium, and high score groups. The performance of LLMs across various subjects and question types is presented in Table~\ref{t3}. We also analyze the relationship between ECS and scoring consistency in Appendix~\ref{a_ecs_ccs}.

We observe that Deepseek-R1 achieves the highest average ECS score, and overall performance generally improves with increasing model parameter size. However, the results also reveal that current LLMs still face significant challenges in maintaining zero-shot consistency for error causes inference. Additionally, model strengths vary notably across different subjects and question types. The lowest average ECS scores are observed on English gap-filling tasks (\textbf{Eng. (G.)}), likely due to the sparse semantics and highly variable answer formats, which make it particularly difficult for LLMs to accurately explain error causes. Compared to scoring tasks, error consistency prediction for short-answer questions in the humanities displays a more polarized pattern: large-scale LLMs achieve relatively high performance, while smaller-scale reasoning-based models tend to make more incorrect inferences. This disparity may be attributed to the lack of explicit alignment between student responses and reference answers in these subjects, which smaller models struggle to resolve due to their limited semantic understanding. Reasoning-based models with fewer than 32B parameters typically perform worse in ECS compared to RLHF-based models. Their more complex reasoning processes often lead to significantly higher inference costs. Overall, ECS effectively highlights the differences between models in terms of consistency in predicting error causes.

\begin{figure}[t!]
  \centering
  \includegraphics[width=\textwidth]{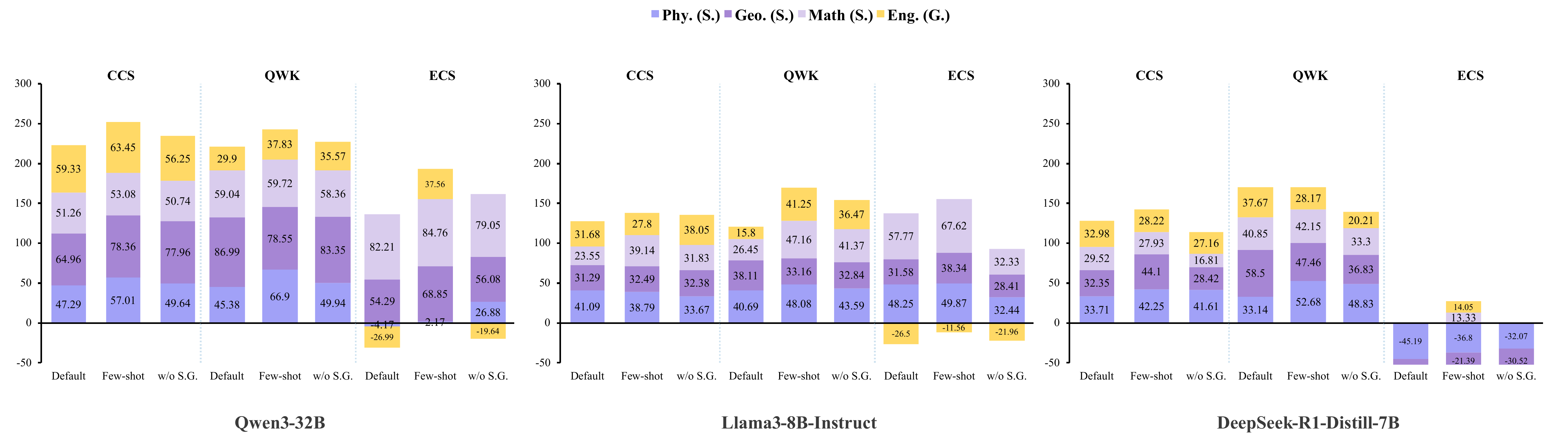}
  \caption{Performance changes of selected models across various subjects under settings of ``few-shot'' and scenarios without scoring guidelines (``w/o S.G.'').}\label{f5}
\end{figure}

\subsection{Impact of Example Demonstrations and Scoring Guidelines}

To evaluate the impact of example samples and scoring guidelines on model performance, we selected underperforming models and corresponding subject datasets based on Tables~\ref{t2} and~\ref{t3}. We then examined performance changes under two conditions: providing 5-shot human-scored examples as demonstrations, and removing the scoring guidelines. The results are illustrated in Figure~\ref{f5}. Overall, the few-shot setting leads to improvements in CCS, QWK, and ECS metrics. However, subject-specific analysis reveals that for some subjects like \textbf{Phy. (S.)} and \textbf{Math (S.)}, few-shot prompting results in decreased CCS scores despite gains in QWK, suggesting that while examples improve overall scoring consistency, the included step-wise labels may mislead the model. Additionally, the absence of scoring guidelines consistently degrades performance across most metrics and tasks, underscoring the importance of providing clear scoring criteria for LLMs in SAS evaluations.

\section{Conclusion}


In this study, we introduce SAS-Bench, the first benchmark specifically designed to evaluate LLMs on SAS tasks. Built from real exam data, it includes 1,030 questions and 4,109 expert-annotated responses with step-wise scores and error causes. SAS-Bench introduces fine-grained metrics that assess model performance across overall and step-wise scoring consistency, as well as explainability based on error cause alignment. Our experiments with various LLMs reveal notable performance gaps among LLMs, highlighting SAS-Bench's value in the development of robust and interpretable automated assessment systems.







\small
\bibliographystyle{IEEEtran}
\bibliography{custom}


\appendix

\section{Limitations}

Ideally, constructing the SAS-bench using real student responses would produce a distribution that closely reflects actual application scenarios. However, the use of LLMs to simulate and generate student responses introduces inherent distributional differences compared to human responses, which constitutes a key limitation of this work. Additionally, obtaining responses across a range of scores in a controlled and balanced manner for each subject is challenging in real-world scenarios. The primary goal of this benchmark is to assess the deviations between LLM evaluations and human assessments in fine-grained response evaluation. Therefore, we emphasize the importance of generating diverse response data and involving professional evaluators for detailed annotations. In the future, we plan to release our annotation system, enabling potential researchers to easily contribute to or expand this work.

\section{Ethics Statement}
Our benchmark datasets utilize LLMs to generate data based on an existing open-source question bank. However, as certain subjects, such as politics and history, may involve content where models can exhibit inherent biases or cognitive distortions, the generated data may reflect social biases present in the pre-training corpus. To mitigate the risk of propagating biased information into scoring models, we recommend that any future use of this dataset for training purposes undergo an additional round of manual review.

\section{Prompts and Examples}\label{a_prompt}

\begin{figure}[h!]
  \centering
  \includegraphics[width=\textwidth]{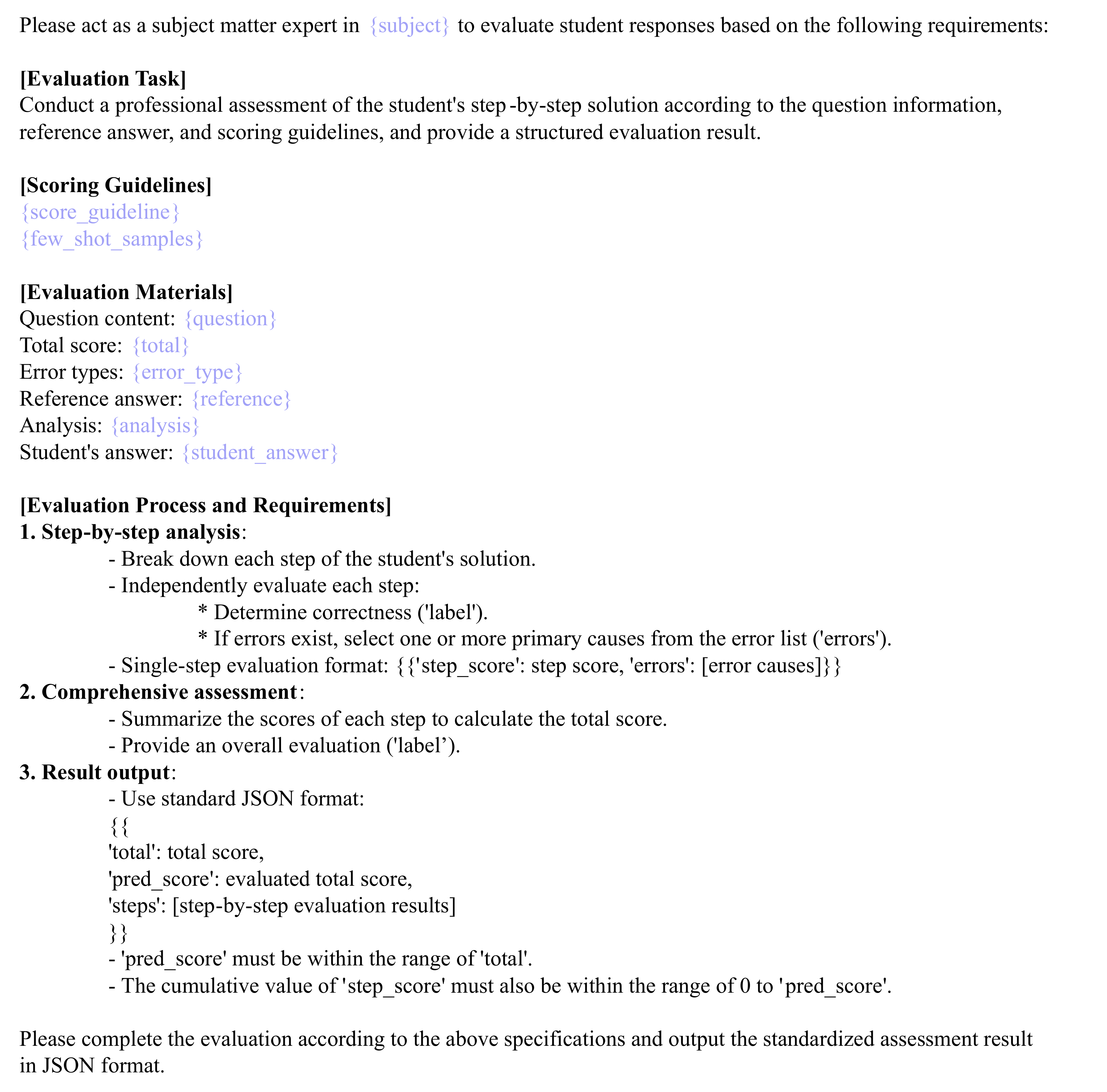}
  \caption{Prompts for model scoring and error cause prediction, where the descriptions are provided in Chinese.}\label{p1}
\end{figure}

\begin{figure}[h!]
  \centering
  \includegraphics[width=\textwidth]{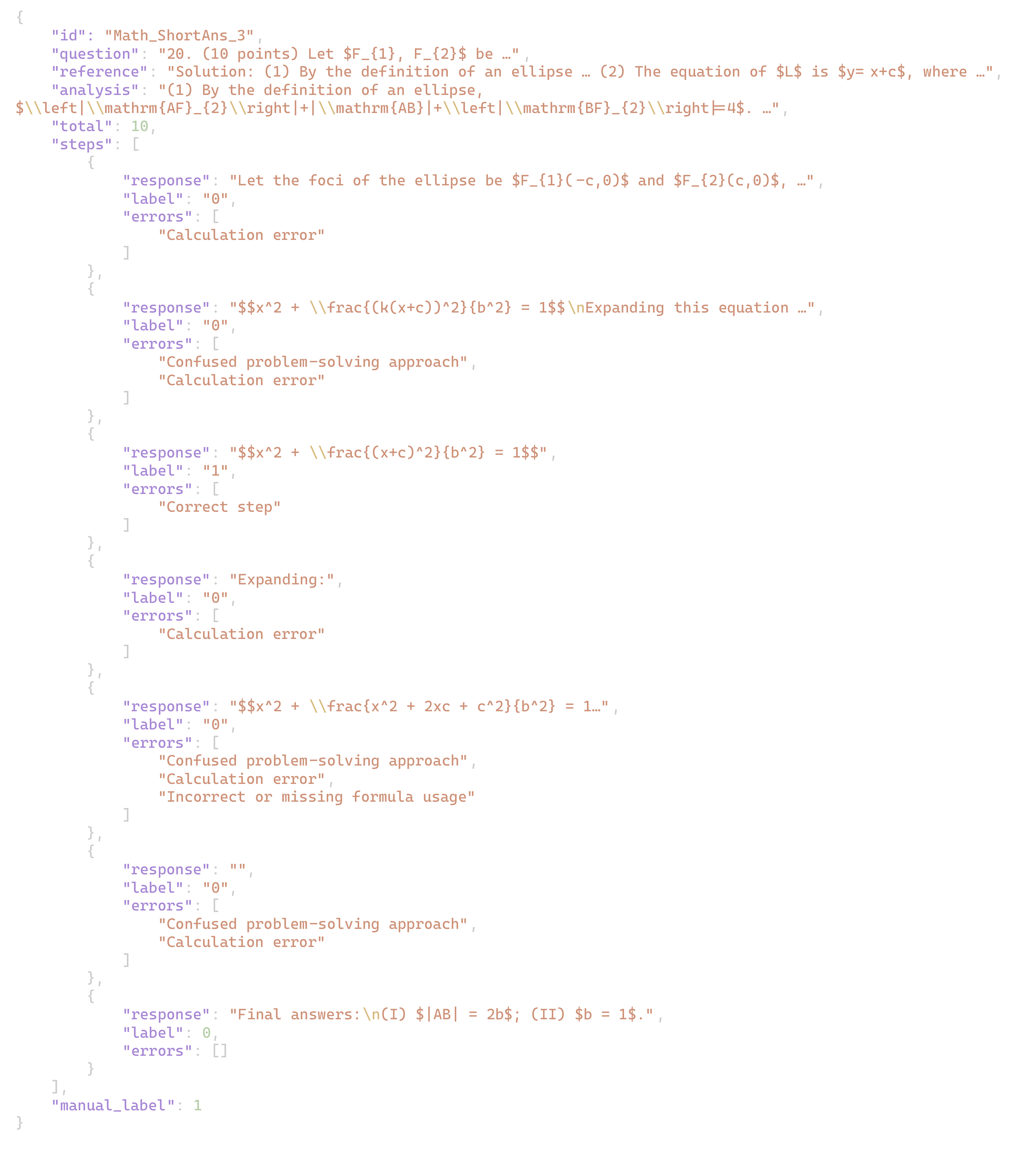}
  \caption{Example format of a math short-answer response annotated with both the score and corresponding error causes.}\label{p2}
\end{figure}

In this section, we present the prompts used for model evaluation in Figure~\ref{p1} and an annotated example of a math short-answer response in Figure~\ref{p2}. Our observations indicate that current LLMs exhibit limitations in numerical reasoning. In particular, when step-wise scores and error causes are positioned before the overall score in the output format, the model tends to produce inflated or biased overall scores. In contrast, generating the overall score (pred\_score) first helps guide and constrain the subsequent step-level scoring and error analysis. Based on this insight, we design the output format to require models to first generate the overall score, followed by the evaluation and analysis of each individual step.

\section{Analysis of Error Cause Detection Capabilities}

\begin{table}[h!]
    \caption{Micro-F1 scores for error causes prediction (\%) across various LLMs, evaluated by subject area and question type. \textbf{S.} indicates Short Answer, \textbf{M.} denotes Multiple Choice, and \textbf{G.} refers to Gap Filling. The best and second-best scores in each category are highlighted in \textbf{bold} and \underline{underlined}, respectively.}
    \label{t4}
    \resizebox{\textwidth}{!}{%
    \begin{tabular}{lccccccccccccc}
    \toprule
    \textbf{Models}        & \textbf{\begin{tabular}[c]{@{}c@{}}Phy. (S.)\end{tabular}} & \textbf{\begin{tabular}[c]{@{}c@{}}Phy. (M.)\end{tabular}} & \textbf{\begin{tabular}[c]{@{}c@{}}His. (S.)\end{tabular}} & \textbf{\begin{tabular}[c]{@{}c@{}}Geo. (S.)\end{tabular}} & \textbf{\begin{tabular}[c]{@{}c@{}}Bio. (G.)\end{tabular}} & \textbf{\begin{tabular}[c]{@{}c@{}}Chi. (G.)\end{tabular}} & \textbf{\begin{tabular}[c]{@{}c@{}}Chi. (S.)\end{tabular}} & \textbf{\begin{tabular}[c]{@{}c@{}}Math (S.)\end{tabular}} & \textbf{\begin{tabular}[c]{@{}c@{}}Math (G.)\end{tabular}} & \textbf{\begin{tabular}[c]{@{}c@{}}Pol. (S.)\end{tabular}} & \textbf{\begin{tabular}[c]{@{}c@{}}Eng. (G.)\end{tabular}} & \textbf{\begin{tabular}[c]{@{}c@{}}Che. (G.)\end{tabular}} & \textbf{Avg.} \\ \midrule
    \multicolumn{14}{c}{Reasoning-based LLMs}                                                                                                                                                                                                                                                                                                                                                                                                                                                                                                                                                                                                                                                                                                                                                                                                                                                      \\ \midrule
Deepseek-R1            & \textbf{51.82}                                                    & 45.71                                                         & 44.47                                                             & 43.12                                                             & {\ul 65.60}                                                        & 83.33                                                                 & 62.35                                                                & \textbf{65.19}                                                    & {\ul 54.67}                                                        & \textbf{60.47}                                                    & 62.46                                                              & 56.07                                                              & \textbf{57.94} \\
QwQ-32B                & 45.81                                                             & 47.73                                                         & 44.30                                                             & 41.38                                                             & 59.19                                                              & 66.07                                                                 & 58.82                                                                & {\ul 64.84}                                                       & 44.63                                                              & {\ul 56.09}                                                       & 58.89                                                              & 53.91                                                              & 53.47          \\
TinyR1-32B-Preview     & 36.20                                                             & 44.44                                                         & {\ul 44.67}                                                       & {\ul 44.53}                                                       & 53.50                                                              & 74.17                                                                 & 58.50                                                                & 49.69                                                             & 37.01                                                              & 53.59                                                             & 52.13                                                              & 48.04                                                              & 49.71          \\
Qwen3-32B              & 46.72                                                             & 47.13                                                         & 41.66                                                             & 40.29                                                             & 59.06                                                              & 81.18                                                                 & 57.34                                                                & 59.53                                                             & 48.98                                                              & 54.57                                                             & 57.61                                                              & 52.21                                                              & 53.86          \\
Qwen3-8B               & 46.35                                                             & \textbf{67.84}                                                & 39.07                                                             & 37.81                                                             & 64.01                                                              & 67.87                                                                 & {\ul 65.09}                                                          & 53.25                                                             & 23.79                                                              & 48.42                                                             & 51.85                                                              & 54.70                                                              & 51.67          \\
MiMo-7B-RL             & 43.42                                                             & 48.45                                                         & 31.49                                                             & 36.79                                                             & 53.09                                                              & 86.21                                                                 & 44.81                                                                & 35.45                                                             & 49.59                                                              & 20.51                                                             & 46.11                                                              & 27.96                                                              & 43.66          \\
Deepseek-Prover-V2-7B  & 38.02                                                             & 35.11                                                         & 27.44                                                             & 27.65                                                             & 48.72                                                              & 23.62                                                                 & 31.36                                                                & 32.91                                                             & 44.96                                                              & 25.00                                                             & 31.41                                                              & 27.55                                                              & 32.81          \\
DeepSeek-R1-Distill-7B & 34.90                                                             & 47.95                                                         & 36.36                                                             & 27.57                                                             & 47.36                                                              & 42.70                                                                 & 29.55                                                                & 26.98                                                             & 51.85                                                              & 38.85                                                             & 50.32                                                              & 19.65                                                              & 37.84          \\ \midrule
\multicolumn{14}{c}{RLHF-based LLMs}                                                                                                                                                                                                                                                                                                                                                                                                                                                                                                                                                                                                                                                                                                                                                                                                                                                           \\ \midrule
Deepseek-V3            & 48.58                                                             & 46.93                                                         & 41.79                                                             & 42.52                                                             & \textbf{67.79}                                                     & \textbf{92.24}                                                        & 62.36                                                                & 63.05                                                             & 53.59                                                              & 52.30                                                             & {\ul 65.60}                                                        & {\ul 56.68}                                                        & {\ul 57.79}    \\
GPT 4o-mini-20240718   & 48.39                                                             & {\ul 48.72}                                                   & \textbf{45.76}                                                    & 40.41                                                             & 54.58                                                              & 59.26                                                                 & 60.60                                                                & 49.13                                                             & 39.90                                                              & 39.14                                                             & 30.59                                                              & 38.65                                                              & 46.26          \\
Llama3.3-70B-Instruct  & {\ul 49.57}                                                       & 45.88                                                         & 42.55                                                             & \textbf{45.54}                                                    & 58.79                                                              & 67.10                                                                 & 53.92                                                                & 51.72                                                             & 43.27                                                              & 47.70                                                             & \textbf{69.64}                                                     & 30.68                                                              & 50.53          \\
Mixtral 8×7B-Instruct  & 33.67                                                             & 43.39                                                         & 41.58                                                             & 37.86                                                             & 52.41                                                              & 33.72                                                                 & 58.86                                                                & 37.57                                                             & 59.28                                                              & 41.66                                                             & 41.53                                                              & 38.58                                                              & 43.34          \\
Qwen2.5-32B-Instruct   & 45.00                                                             & 45.71                                                         & 40.75                                                             & 41.75                                                             & 62.60                                                              & {\ul 89.63}                                                           & 54.27                                                                & 61.39                                                             & 42.67                                                              & 53.85                                                             & 48.37                                                              & \textbf{56.77}                                                     & 53.56          \\
Qwen2.5-14B-Instruct   & 43.70                                                             & 45.24                                                         & 39.86                                                             & 36.82                                                             & 61.72                                                              & 88.26                                                                 & \textbf{65.34}                                                       & 48.14                                                             & 49.49                                                              & 49.17                                                             & 54.75                                                              & 50.43                                                              & 52.74          \\
GLM4-9B-Chat           & 40.00                                                             & 44.32                                                         & 41.10                                                             & 37.14                                                             & 55.79                                                              & 48.31                                                                 & 57.48                                                                & 38.76                                                             & 38.73                                                              & 42.35                                                             & 41.50                                                              & 42.42                                                              & 43.99          \\
Llama3-8B-Instruct     & 42.95                                                             & 37.63                                                         & 34.15                                                             & 32.64                                                             & 56.28                                                              & 50.30                                                                 & 53.28                                                                & 47.60                                                             & \textbf{68.76}                                                     & 24.89                                                             & 47.11                                                              & 46.41                                                              & 45.17          \\ \bottomrule
\end{tabular}%
    }
\end{table}

To assess whether models can accurately identify all relevant error causes in student responses, we evaluate their detection capabilities using the Micro-F1 score across different subjects and question types. For each question, we first merge and deduplicate the step-wise error causes predicted by the model and those annotated by human experts. We then compute true positives (TP), false positives (FP), and false negatives (FN) by checking the presence of each error cause in the predicted and reference sets. The Micro-F1 score is calculated based on the aggregated TP, FP, and FN across all questions. The results are summarized in Table~\ref{t4}.

Deepseek-R1 still achieves the highest average F1 score. However, compared to the ECS results in Table~\ref{t3}, all models demonstrate markedly higher F1 scores and narrower performance gaps. This suggests that while current LLMs are generally capable of identifying the overall set of error causes in a response, they still face challenges in accurately reasoning through step-wise, fine-grained errors when compared to human annotations. The comparison between F1 and ECS further underscores the value of ECS as a metric. Unlike F1, which primarily reflects aggregate detection performance, ECS captures the consistency of reasoning across steps, offering a more nuanced and informative measure of model explainability in SAS tasks.

\section{Analyzing the Relationship Between ECS and Scoring Performance}\label{a_ecs_ccs}

\begin{figure}[h!]
  \centering
  \includegraphics[width=\textwidth]{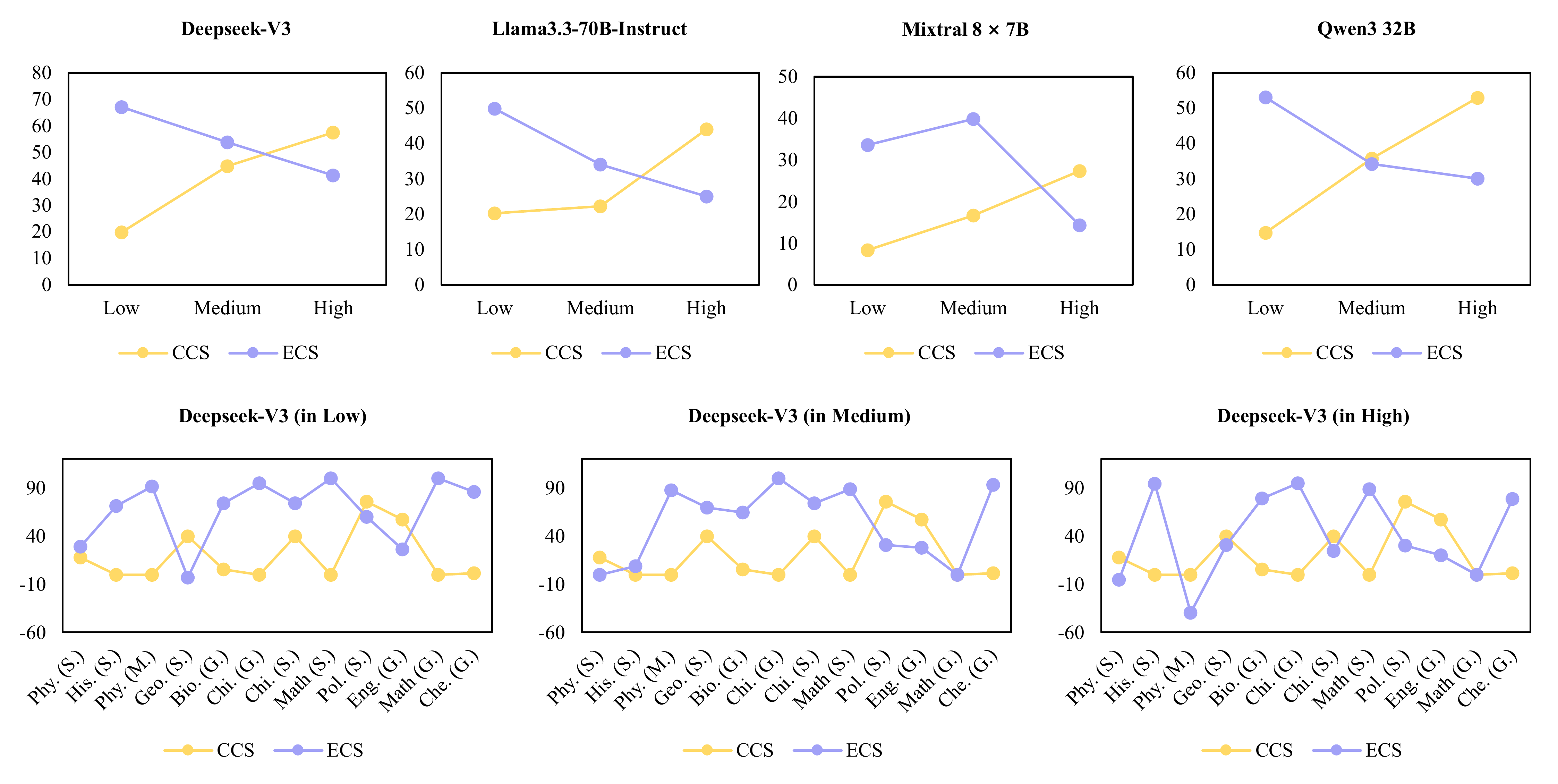}
  \caption{ECS and CCS comparison across low, medium, and high score intervals. The upper line chart illustrates the average ECS and CCS scores of different models across all subjects and question types, while the lower chart provides a detailed breakdown of Deepseek-V3's performance across individual subject areas and question types.}\label{f6}
\end{figure}

To further investigate the relationship between ECS and model scoring performance, we analyze the evaluation results of four representative models: Deepseek-V3, LLaMA3.3-70B-Instruct, Mixtral 8×7B, and Qwen3-32B. Following the score-based partitioning strategy described in $\S$\ref{s4.3}, we divide the evaluation samples into three intervals: low, medium, and high based on the overall scores assigned by human annotators. For each interval, we compute the average CCS and ECS across all subjects and question types. The results are presented in Figure~\ref{f6}.

Notably, we observe an overall inverse trend between ECS and CCS: as ECS decreases, CCS tends to increase, and vice versa. To explore this phenomenon in more detail, we further examine Deepseek-V3's ECS and CCS across different intervals for each subject and question type. This finer-grained analysis confirms the inverse relationship, suggesting that a model's consistency in error cause identification may affect its overall scoring performance. Moreover, ECS values tend to drop more significantly in the high-score interval, indicating that models are more effective at identifying error causes in lower-scoring responses. One possible explanation is that models may over-focus on local error causes, leading to reduced leniency compared to human annotators and ultimately resulting in lower scoring consistency. When considered alongside the results in Table~\ref{t4}, which show that models achieve higher performance in overall error cause detection (as measured by Micro-F1) than in ECS, these findings suggest that the relatively low ECS scores are primarily attributable to the models' challenges in accurately reasoning about step-wise error causes in medium- and high-scoring responses.

\begin{figure}[t!]
  \centering
  \includegraphics[width=\textwidth]{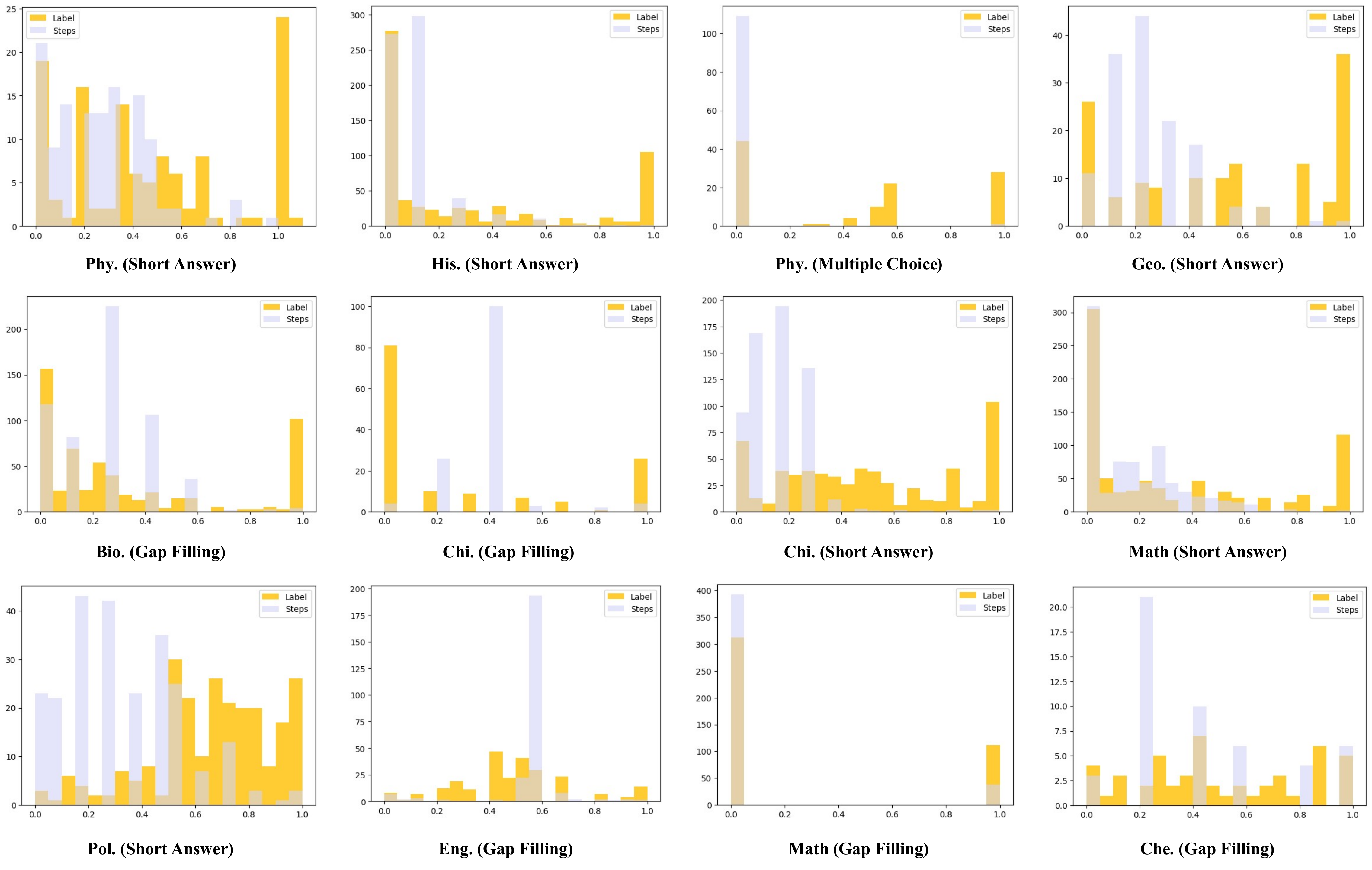}
  \caption{Normalized distributions of total human-annotated scores and the number of steps in the dataset.}\label{f7}
\end{figure}

\section{Data Distribution Visualization}

To better understand the composition of our dataset, we visualize the distributions of human-annotated overall scores and the number of reasoning steps in student responses across different subject areas and question types, as shown in Figure~\ref{f7}. For clarity, both scores and step counts are normalized.

As the figure illustrates, all subject-question types, except for \textbf{Math (G.)}, which consists of single-step gap-filling questions and thus only includes full or zero scores, span a broad range of scores across low, medium, and high intervals. While real-world student response distributions typically approximate a normal distribution, our dataset contains a larger proportion of responses with either full marks or zero scores. This is largely due to the data selection process. However, as described in $\S$~\ref{s3.2}, we applied rigorous data cleaning to ensure high semantic diversity within each score range. This diversity is crucial for evaluating whether models can consistently and accurately score varied responses that fall within the same scoring band.

\begin{figure}[t!]
  \centering
  \includegraphics[width=\textwidth]{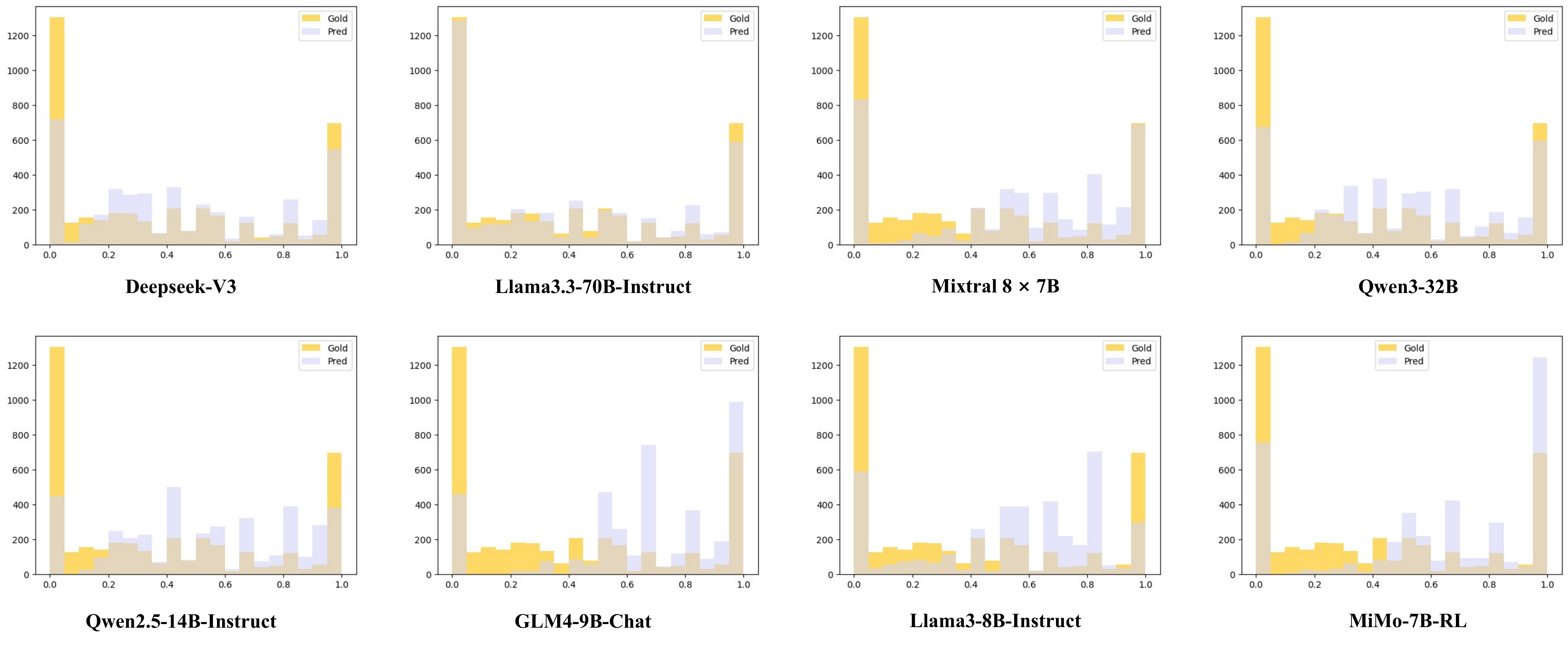}
  \caption{Normalized distributions of total human-annotated scores and the number of steps in the dataset.}\label{f8}
\end{figure}

\section{Analysis of Predicted Score Distributions}\label{a_pred}

We selected eight representative models of varying sizes and compared their predicted overall score distributions with human-annotated scores across the entire dataset, as illustrated in Figure~\ref{f8}.

Overall, model predictions tend to concentrate in the mid-score range, reflecting a more conservative scoring pattern compared to human annotators, models are generally less inclined to assign either full or zero scores. Notably, we also observe considerable variation across models in their scoring tendencies within this middle range. Among them, LLaMA3.3-70B-Instruct shows a score distribution most closely resembling that of human annotations. However, when we examine the step-wise evaluation metrics presented in Table~\ref{t2} and Table~\ref{t3}, LLaMA3.3-70B-Instruct still lags behind Deepseek-V3 in both CCS and ECS performance. This discrepancy underscores the strength of our proposed evaluation metrics, which capture deeper aspects of model performance in SAS tasks, beyond mere alignment with overall scores.



\end{document}